\documentclass[runningheads]{llncs}

\usepackage{amsmath,amsfonts,bm}

\def\eqref#1{equation~\ref{#1}}

\def\1{\bm{1}}

\DeclareMathAlphabet{\mathsfit}{\encodingdefault}{\sfdefault}{m}{sl}
\SetMathAlphabet{\mathsfit}{bold}{\encodingdefault}{\sfdefault}{bx}{n}

\newcommand{\R}{\mathbb{R}}

\usepackage[T1]{fontenc}
\usepackage{hyperref} 
\usepackage{color}

\usepackage{orcidlink}

\newcommand{\wtilde}[1]{\overset{\sim}{#1}}

\newcommand{\sub}[1]{\textsubscript{\texttt{[#1]}}}

\usepackage{comment}
\usepackage{graphicx}
\usepackage{amsmath}
\usepackage{amssymb}
\usepackage{tabularx}
\usepackage{booktabs} 
\usepackage{siunitx}
\usepackage{xcolor}
\usepackage{comment}
\usepackage[capitalize]{cleveref}
\crefname{section}{Sec.}{Secs.}
\Crefname{section}{Section}{Sections}
\Crefname{table}{Table}{Tables}
\crefname{table}{Tab.}{Tabs.}

\usepackage[table]{xcolor}    %
\usepackage{booktabs}         %
\usepackage{siunitx}          %
\usepackage{caption}          %
\usepackage{adjustbox}        %
\usepackage{cases}
\usepackage{colortbl}
\usepackage{multirow}
\usepackage{bm}
\usepackage{subcaption}

\begin{document}
\title{Accelerating Vision Foundation Models with Drop-in Depthwise Convolution}

\author{Carmelo Scribano\inst{1\thanks{Work done while visiting INSAIT}}\orcidlink{0000-0003-1006-7826}, Mohammad Mahdi\inst{2}, Nedyalko Prisadnikov\inst{2}\orcidlink{0009-0004-2202-8431}, Yuqian Fu\inst{2}\orcidlink{0000-0002-0412-5500}, Giorgia Franchini\inst{1}\orcidlink{0000-0001-9082-8087}, Danda Pani Paudel\inst{2}\orcidlink{0000-0002-1739-1867}, Marko Bertogna\inst{1}\orcidlink{0000-0003-2115-4853}, Luc Van Gool\inst{2}\orcidlink{0000-0002-3445-5711}}

\authorrunning{C. Scribano et al.}

\institute{University of Modena and Reggio Emilia, Modena, Italy. \\
\email{\{name.surname\}@unimore.it} \and
INSAIT, Sofia University ``St. Kliment Ohridski'', Sofia, Bulgaria\\
\email{\{name.surname\}@insait.ai}
}

\maketitle              %
\setcounter{footnote}{0}
\begin{abstract}
Pretrained vision foundation models deliver strong performance across tasks with limited fine-tuning. However, their Vision Transformer (ViT) backbones impose high inference costs, limiting deployment on resource-constrained devices. In this work, we accelerate large-scale pretrained ViTs while preserving their feature extraction capabilities by exploiting the intrinsic convolution-like behavior of some attention heads. Specifically, we introduce an efficient depthwise convolution-based layer that serves as a drop-in replacement for these heads. Additionally, we propose simple strategies to identify which heads can be replaced and introduce a fine-tuning procedure that recovers downstream task performance. Across both image classification and segmentation tasks, our method achieves 17–20\% inference speedup with minimal performance degradation. We validate the approach through detailed derivations, extensive experiments, and efficiency benchmarks. The reference implementation is publicly available\footnote{\url{https://github.com/cscribano/DWConv\_VFM}}. %

\end{abstract}
\section{Introduction}
Large-scale-pretrained Vision Transformers (ViTs) (\cite{dosovitskiy2021an}) have emerged as a powerful paradigm in modern machine learning. With minimal fine-tuning, they can perform competitively on a wide range of downstream vision tasks. Dominating paradigms include DINO (\cite{caron2021emerging,oquab2023dinov2}), MAE (\cite{he2022masked}), and CLIP (\cite{radford2021learning}). A noticeable downside is the high inference cost of the ViT architecture,  especially when targeting inference on low-power edge devices. Although various methods have been proposed to reduce the inference cost of ViTs, several challenges remain. Popular approaches leverage a reduction in token counts (\cite{bolya2022tome,meng2022adavit,graham2021levit}), which disrupts the spatial structure of the features and limits the applicability to dense prediction tasks (i.e, segmentation). Another common occurrence is the reliance on dynamic conditional execution flow or dynamically shaped tensors, seldom supported by high-performance inference frameworks. In addition, for finetuned foundation models, it is fundamental to preserve the large-scale pretrained weights, which is usually not the case in efficient transformer backbones (\cite{cai2023efficientvit,mehta2022mobilevit,wadekar2022mobilevitv3}).

\begin{figure}[h!]
    \centering
    \begin{subfigure}[b]{0.48\textwidth}
        \centering
        \includegraphics[width=\textwidth,clip]{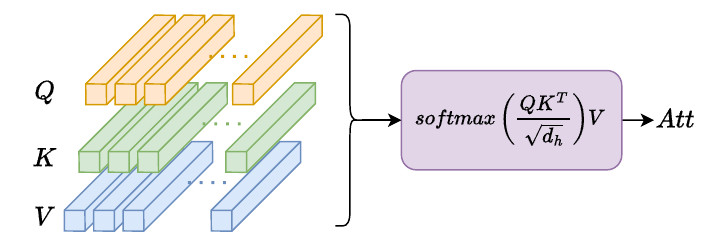}
        \caption{Dot-product Attention}
    \end{subfigure}
    \begin{subfigure}[b]{0.48\textwidth}
        \centering
        \includegraphics[width=\textwidth,clip]{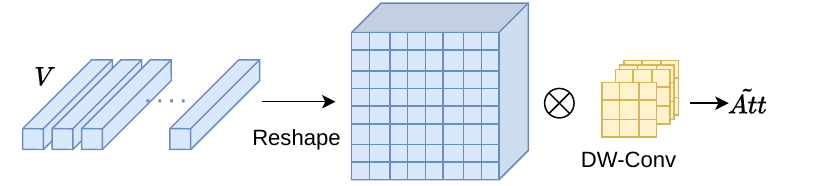}
        \vspace{0.5em} %
        \caption{Proposed Drop-in replacement}
    \end{subfigure}
    \caption{\textbf{Illustration of the proposed drop-in approximation.} 
    We replace attention (a) with a Depthwise convolution (b), which improves inference speed while reusing the pre-trained network parameters for performance.}
    \label{fig:dwlayer}
\end{figure}

\noindent To address these challenges, in this paper, we propose an efficient,\emph{ drop-in acceleration method for foundation ViTs}. Building on previous research (\Cref{sec:rel_convatt}), we assume that several Multi-head Self Attention (MhSA) \cite{vaswani2017attention} heads learn highly localized, static patterns, with a structure closely resembling convolution. We propose to replace a carefully selected set of computationally intensive Self-Attention operators with a much more efficient depthwise convolution over the reshaped Value tensors (\Cref{fig:dwlayer}). Our method works as a drop-in replacement, recovering the performance of the large-scale pretrained model with limited fine-tuning, showing minimal performance loss while achieving 17\% to over 20\% speedup.

\noindent The contributions of this work can be summarized as follows.
\begin{itemize}   

    \item We derive an efficient formulation that serves as a drop-in replacement for attention heads that learn particular convolution-like structures. We later show that head ensembling from \cite{he2024pruning} can be made explicit and generalized to our setting, benefiting from the more efficient formulation.
    
    \item We propose a simple methodology to identify the heads to be approximated by convolution. We also validate the proposed heuristic against a more sophisticated solution that we reformulate for this context.

    \item We explicitly consider the problem of performance optimization in the context of pretrained foundation models, targeting a realistic deployment scenario, focusing on a popular edge platform device (Nvidia Orin Nano) and an appropriate inference framework (TensorRT). %
    
\end{itemize}

\noindent We believe that the work presented addresses numerous challenges that have been only partially covered in the literature. It also allows for casting the performance improvement of vision foundation models from a different perspective, which in the future might become part of more advanced pruning frameworks. 

\section{Related Work}
\label{sec:rel}

\subsection{Pruning of Transformers}
\label{subsec:pruning}

The idea of reducing the complexity of a neural network by eliminating less important parameters, connections, and layers dates back to the early deep learning era (\cite{lecun1989optimal,hassibi1992second}). Similar techniques, while applicable to BERT \cite{devlin2019bert} style transformers (\cite{sanh2020movement,chen2020lottery,brix2020successfully}), induce \textit{unstructured} sparsity, often causing overhead due to irregular memory access. Some works partially address this via block sparsity (\cite{lagunas2021block,xu2024lpvit}). Structured pruning of attention heads was explored in \cite{michel2019sixteen}, with \cite{voita2019analyzing} introducing stochastic gating to select heads during training. Building on this, \cite{behnke2020losing} identifies prunable heads early using confidence scores, while DSP \cite{li2021differentiable} introduces explicit control over pruning ratios.

\noindent Some pruning methods for ViTs require access to training (\cite{prasetyo4529273sparse,lin2024mlp}), others rely on distillation (\cite{yu2022unified,yang2023global}). Many focus on pretrained DeiT models (\cite{pmlr-v139-touvron21a}) for classification (\cite{zheng2022savit,yu2022unified,he2024pruning,lin2024mlp}). SPViT \cite{he2024pruning} prune MhSA blocks during fine-tuning into learned convolutional layers formulated under the sufficient condition of \cite{Cordonnier2020On}. That assumption implicitly collapses the block into a single convolution. In contrast, we derive a more general formulation applicable to pretrained foundation models, and later show how SPViT arises as a special case within our framework. Lambda-ViT (\cite{lin2024mlp}) gradually degenerates MhSA blocks to identity mappings, guided by a transfer-entropy measure. Both target DeiT backbones for classification tasks.

\subsection{Attention Modeling as Convolution}\label{sec:rel_convatt}
Several works have investigated the similarity between convolution and spatial relationships learned by attention. (\cite{raghu2021vision}) suggests substantial differences in learned patterns, though some shallow-layer heads focus on local features. The ability of attention to capture localized patterns is further explored in \cite{jelassi2022vision}, emphasizing the role of positional encoding in learning spatial connectivity. \cite{Cordonnier2020On} constructively proves that, under strong assumptions, an MhSA block can implement a convolutional layer if each attention head attends to a distinct location within a region the size of a convolutional kernel. In practice, as stated by the authors, this stands only as a \emph{sufficient} condition. We further discuss this in \Cref{subsec:formul}. An influential contribution to our work \cite{han2022on} discusses the relationship between attention (local) and Depth-Wise (DW) convolution, highlighting key properties that we further develop in \Cref{subsec:sigmacrit}. 

\subsection{Other Approaches}

\textit{Token reduction methods} \cite{rao2021dynamicvit,yin2022vit,marin2021token,bolya2022tome} aim to reduce the number of tokens processed by transformers via removal or aggregation. These methods face limitations: token reduction disrupts spatial structure, making them unsuitable for dense prediction tasks like segmentation or depth estimation. Moreover, techniques relying on clustering, gather/scatter operations, or dynamic shapes often incur significant overhead on specialized inference frameworks. \textit{Efficient Attentions} contributions propose alternative formulations of the attention layer to mitigate computational complexity (\cite{shen2021efficient,xiong2021nystromformer,yao2024mobile}). These methods are generally not intended to serve as drop-in replacements in pretrained Vision Transformers, where preserving the pretrained weights and model behavior is essential and retraining from scratch is impractical. Alternative research directions include focusing on the algorithmic and implementation refinement of more intensive operations to take full advantage of hardware capabilities, a prominent example is Flash-Attention (\cite{dao2022flashattention,dao2023flashattention2}).

\section{Methodology}
\label{sec:method}
\subsection{Background}

\paragraph{Vision Transformers and MhSA.}
The ViT input is defined by splitting an image $I \in \R^{H \times W \times 3}$ into non-overlapping patches of size $p\times p$, flattening each patch into a vector in $\R^{3p^2}$ and projecting it to a $d$-dimensional embedding. Assuming $\frac{H}{p} = \frac{W}{p} = \sqrt{n}$, this yields an input tensor $X \in \R^{n \times d}$ of $n$ patch embeddings, to which is added a positional encoding. Most implementations, including Dino-V2, prepend a $\texttt{[cls]}$ token, which we ignore for now. The sequence is processed through transformer blocks alternating Multi-head Self-Attention (MhSA) and feed-forward layers.

Let $n_h$ be the number of attention heads, the MhSA is parametrized by $W^Q, W^K,W^V\in\R^{d_i\times (n_h*d_h)}$ and $W^O\in\R^{(d_h * n_h)\times d_o}$ where typically $d_i=d_o=d$ and $d_h * n_h = d$, as we assume from now on. Defining Query, Keys, and Values for the $h$-th head as:
\begin{equation}
    Q^h=XW^Q_{[:,h,:]} K^h=XW^K_{[:,h,:]} V^h=XW^V_{[:,h,:]}
    \label{eq:qkv}
\end{equation}

For the \textit{h}-th head, the attention is computed as:
\begin{gather}
    Att(X)^h = E(X)^hV^{h} \label{eq:atnhead} \\
    E(X)^h = softmax\left(\frac{Q^h{K^{h}}^\top}{\sqrt{d_h}}\right)  \label{eq:energy}
\end{gather}
where $E(X)^h\in\R^{n\times n}$ is the attention (energy) matrix. Concatenating the head outputs yields the full MhSA output:
\begin{equation}
    MhSA(X) = [Att^1(X)\,\|\,\dots\,\|\,Att^{n_h}(X)]W^O
    \label{eq:mhsa}
\end{equation}
with $\|\,$ denoting concatenation along the embedding dimension.

\paragraph{Convolutional Layers.}

A Convolutional layer with kernel size $k$ is parametrized by weights  $W^C\in R^{k\times k \times c_i \times c_o}$. 
Expliciting the symmetric shift set of size $k\times k$ as $\Delta_k = \{(s,r)\in \mathbb{Z}^2 : -\lfloor k/2 \rfloor\leq s,r\leq\lfloor k/2 \rfloor\}$ (assuming a stride factor of $1$), the output at location $(i,j)$ for an input $X\in R^{h\times w \times c_i}$ is defined as:
\begin{equation}
    Conv(X, W^C)_{i,j} = \sum_{r,s\in\Delta_k}{W^C_{r,s}}^\top X_{i+r, j+s}
\end{equation}
\noindent Thus, convolution produces a weighted local aggregation of the $k\times k$ neighborhood centered at $(i,j)$. Similarly, depth-wise convolution performs local aggregation by applying a distinct spatial filter to each channel independently. With $W^D\in\R^{k\times k\times c_i}$, we write
\begin{equation}
    Conv_{DW}(X,W^D)_{i,j} = \sum_{(r,s)\in\Delta_k}  W^D_{r,s} \odot X_{i+r,\,j+s}
\end{equation}
where $\odot$ denotes elementwise multiplication.\\

\subsection{Convolutional Approximation}
\label{subsec:formul}

In this section, we formalize how attention can be approximated by convolution, and present the efficient depthwise decomposition at the core of our method. We later show that the SPViT \cite{he2024pruning} bottleneck block arises as a special case of our formulation, extended with head ensembling.

\subsubsection{Drop-in Depthwise Formulation.}
\label{subsec:csa}

Consider a single attention head as in \Cref{eq:atnhead}. For clarity, we reshape the input $X\in\R^{n\times d}$ to $X\in\R^{m\times m\times d}$ with $m=\sqrt{n}$, recovering spatial structure. Accordingly, $E(X)^h\in\R^{n\times n}$ can be viewed as $E(X)^h\in\R^{(m\times m)\times(m\times m)}$, and the values as $V^h\in\R^{m\times m\times d_h}$. When writing two-dimensional indices, we always refer to these unflattened tensors.
The explicit form of attention at location $(i,j)$ is:
\begin{equation}
    Att(X)^h_{i,j} = \sum_{r,s\in\Delta_{m}} E^h(X)_{(i,j),(i+r,j+s)}V^h_{i+r,j+s}
    \label{eq:att_explicit}
\end{equation}
where $\Delta_m$ denotes the full receptive field that spans the whole $E(X)^h$. This resembles convolution, with key differences: the weights $E^h(X)_{(i,j),(i+r,j+s)}$ for spatial aggregation depend on both the input $X$ and the query position $(i,j)$,  and convolutional kernels are fixed parameters shared across locations.  

We approximate attention by assuming that some heads can be replaced by \emph{input-independent kernels} restricted to a local neighborhood  $\Delta_k\subset\Delta_m$. Formally, for head $h$ we write:
\begin{equation}
    Att(X)^h_{i,j}
    \;\approx\; \sum_{(r,s)\in\Delta_k} K^h_{r,s}\, V^h_{i+r,\,j+s}
    \label{eq:att_aprox}
\end{equation}
where $K^h\in\R^{k\times k}$ are trainable parameters learned during fine-tuning. %

\paragraph{Full convolution formulation.}
A direct implementation of \Cref{eq:att_aprox} is to fold $K^h$ into the value projection $W^V$,
producing a kernel $W^{Vh}\in\R^{k\times k \times d_i \times d_h}$:
\begin{gather}
    \wtilde{Att}_{C}(X)^h = Conv(X, W^{Vh}) \label{eq:attconv_full}\\
    W^{Vh}_{r,s} = K^h_{r,s}\,W^V_{[:,h,:]}, \quad (r,s)\in\Delta_k
    \label{eq:full_kernel}
\end{gather}
This formulation, while being a faithful analogue of \cref{eq:att_aprox}, is not appealing from a complexity standpoint, as assessed in \Cref{sec:convatt_results}.%

\paragraph{Depthwise decomposition.}
To reduce cost, we separate the pointwise value projection from the spatial aggregation. We first compute values $V^h=XW^V_{[:,h,:]}$, then apply a
depthwise convolution with head-specific kernels $\vec{K}^h\in\R^{k\times k\times 1\times d_h}$:
\begin{equation}
    \boxed{\wtilde{Att}_{DW}(X)^h = Conv_{DW}(V^h, \vec{K}^h)}
    \label{eq:attconv_dwpw}
\end{equation}
Compared to the full convolution, the complexity is reduced from $O(k^2 d_i d_h)$ to $O(d_i d_h + k^2 d_h)$.  In the full formulation (\Cref{eq:attconv_full}), each $K^h_{r,s}$ is shared across all $d_h$ channels, enforcing a single spatial pattern. Depthwise kernels $\vec{K}^h$, instead, provide one $k\times k$ filter per channel, enabling distinct spatial aggregations. While channel-wise sharing could be easily implemented, we relax it without affecting performance.  This layer can replace any subset of attention heads in the MhSA block (\cref{eq:mhsa}). Its implementation is schematized in \Cref{fig:dwlayer} and thoroughly evaluated in the experimental section.

\paragraph{Head-ensembling.}
The formulation of \cite{he2024pruning} builds on the sufficient condition of \cite{Cordonnier2020On}, where each head attends to a distinct spatial location within a local neighborhood. This assumption implicitly enforces a degenerate head ensembling, causing the MhSA block to collapse into a single effective head. In contrast, we derive the ensembling explicitly and show that it extends beyond the full convolution setting. Specifically, by assigning learnable combination weights $\gamma \in \mathbb{R}^{n_h}$ to control the contribution of each head, the ensembled value and output projections become:
\begin{gather}
    W^{Ve} = \sum_{h=1}^{n_h} \sigma(\gamma_h)\, W^V_{[:,h,:]}, \qquad
    W^{Oe} = \sum_{h=1}^{n_h} \sigma(\gamma_h)\, W^O_{[:,h,:]},
\end{gather}
with $\sigma(\cdot)$ a softmax over heads. Crucially, this explicit ensembling extends naturally to our depthwise formulation:
\begin{equation}
    \begin{gathered}
    \wtilde{MhSA}^e_{DW}(X) = Conv_{DW}(V^e,\vec{K}^e)W^{Oe} \\
    s.t \quad V^e = XW^{Ve}\\
\end{gathered}
\label{eq:mhsa_dw_ensemble}
\end{equation}
and $\vec{K}^e$ denotes the depthwise convolution kernel as in \Cref{eq:attconv_dwpw}. In this view, \cite{he2024pruning} arises as a \emph{special case} of our framework. We henceforth distinguish between the \underline{ensembled} formulation (\Cref{eq:mhsa_dw_ensemble}) and the \underline{unensembled} formulation (direct head replacement via \Cref{eq:attconv_dwpw}).

\subsection{Layer Selection}
\label{subsec:stc_crit}
Given a target of $p_h$ heads to approximate with convolution, selection can be either \underline{scattered}, replacing arbitrary heads across the model, or \underline{blockwise}, replacing all $n_h$ heads within $p_b$ MhSA blocks. As shown in \cref{subsec:results_selection}, the blockwise strategy yields higher inference efficiency, and we therefore adopt it as the default. Below, we introduce two criteria for defining the head set $\mathcal{S}$, both applicable to either selection mode.

\subsubsection{Proposed Criterion.}
\label{subsec:sigmacrit}

As briefly mentioned, the approximation for $E(X)^h$ introduced in \Cref{eq:att_aprox} is equivalent to real attention under the conditions of Locality (L), translation invariance (TI), and Input Invariance (II). For a receptive field $\Delta_k$ and displacement $(s,r)\in\Delta_k$, these are:
\begin{subnumcases}{}
 \text{(L)}: E(X)^h_{(i,j),(u,v)} \neq 0 \;\; \text{only if } (u-i, v-j) \in \Delta_k \label{eq:suba}\\
 \text{(TI)}: E(X)^h_{(i,j),(i+s,j+r)} = E(X)^h_{(l,t),(l+s,t+r)} \quad \forall (i,j),(l,t) \label{eq:subb}\\ 
 \text{(II)}: E(X)^h = E(Y)^h \quad \forall X,Y \label{eq:subc}
\end{subnumcases}

\paragraph{Criterion Definition} We empirically establish the sum of the pointwise standard deviation of $E(X)^h$ as a simple and effective proxy for identifying convolutional-like heads. Concretely, for each head $h\in\{1,\dots,N_h\}$, where $N_h=n_hn_b$ is the cumulative number of heads across $n_b$ blocks, we compute the pointwise standard deviation $\sigma_{E^h}$ of $E(X)^h$ over $N_s$ input samples. Direct computation of $\sigma_{E^h}$ impractical, since accumulating $E^h(X_i)$ for a reasonable input set (i.e., $N_s=1000$) would require over 500GB of memory. We use Welford’s algorithm \cite{welford1962note} to compute $\sigma_{E^h}$ online in a single pass. 
We then define a scalar score
\begin{equation}
    \Sigma_h = \sum \sigma_{E^h},
    \label{eq:sigma_h}
\end{equation}
summing over all entries of $E^h$. 
We select as candidate heads the set $\mathcal{S}_h^{[p_h]}$ of the $p_h$ heads with the smallest $\Sigma_h$. 
\noindent In the blockwise setting, we adopt the same criterion at block level. 
For each block $b\in\{1,\dots,n_b\}$ we compute the mean score across its $n_h$ heads:
\begin{equation}
    \Sigma_b = \frac{1}{n_h}\sum_{h\in[n_h]}\Sigma_h,
\end{equation}
and select the set $\mathcal{S}_b^{[p_b]}$ of $p_b$ blocks with the smallest $\Sigma_b$.

\paragraph{Rationale}
By construction, $\Sigma_h=0$ is both necessary and sufficient for the Input-Invariance property \Cref{eq:subc}. In the limit $\Sigma_h \to 0$, the kernel collapses to its expectation $E^h(X)\to\mu_{E^h}$, eliminating dependence on the input. Our heuristic is driven by the observation that Input-Invariance is the most stringent property: once achieved, it forces the head to ignore input variation entirely and reduces to a positional-only operator. In this regime, $E^h(X)$ derives solely from the learned positional encodings shared across heads in standard ViTs. Positional attention mechanisms have been linked to spatial connectivity patterns (patch association; \cite{jelassi2022vision}), capturing the locality and translation-like structure that underlies convolutional inductive biases. We therefore use $\Sigma_h$ as a heuristic, motivated by the Input-Invariance principle, but ultimately empirical.

\subsubsection{Stochastic Gating.}\label{subsec:dsp}
As an alternative to the presented criterion, we propose a comparison with a selection method derived from Differentiable Subset Pruning (DSP) \cite{li2021differentiable}. While DSP in origin prunes transformer heads, we can easily generalize it to our scope. For simplicity, we present this mechanism in the blockwise case, although it can be trivially extended to the scattered selection. We define a set of trainable parameters $w^b\in \R^{n_b}$, leveraging the $\text{topk}(\cdot)$ operator we can retrieve the largest $p_b$ elements of $w^b$:
\begin{equation}
    \text{topk}(w^b, p_b)_i = \begin{cases}
    1  &\text{if}\quad i\in \mathcal{S}_b^{[p_b]}\\
    0 &\text{otherwise}\\
\end{cases}
\end{equation}
Defining $\overline{w}^b_i=\text{topk}(w^b, p_b)_i$, we can implement a simple gating mechanism to select the chosen operation during the forward pass:
\begin{equation}
        \begin{aligned}
        \overline{MhSA}^i(X) = (1-\overline{w}^b_i)MhSA(X)^i & + \overline{w}^b_i\wtilde{MhSA}(X)^i
        \end{aligned}
        \label{eq:gumbel_mhsa}
\end{equation}
This formulation involves only a minimal increase in the number of parameters and no additional loss terms. Since the $\text{topk}$ operator is non-differentiable, to learn $w^b$ during training, the Gumbel top-k relaxation \cite{pmlr-v97-kool19a} is used, an extension of the Gumbel softmax trick \cite{jang2016categorical}, which provides a differentiable approximation $\tilde{w}^b$ of the hard selection $\overline{w}^b$, controlled by a temperature $\tau$. As $\tau \to 0$, $\tilde{w}^b$ approaches $\overline{w}^b$; we follow the $\tau$ annealing schedule from DSP authors. %

\section{Experiments and Comparisons}
\label{sec:exp}

\subsection{Experimental Setup}

Unless otherwise specified, our analysis is based on the Dino-V2 (\cite{oquab2023dinov2}) model fine-tuned on various downstream tasks. This choice simplifies the discussion of the results. In \Cref{sec:convatt_results}, we show that similar results hold when using other vision foundation transformer backbones.

\noindent\textit{Benchmarking.} To reflect real deployment conditions, we mainly target TensorRT on Nvidia Jetson Orin Nano to profile inference performances. Unlike general-purpose frameworks (e.g., PyTorch), TensorRT compiles models into optimized GPU kernels, highlighting real-world performance constraints. \\
\noindent\textit{Task Performance.} We evaluate fine-tuned models on semantic segmentation (COCO~\cite{cocodataset}, ADE20K~\cite{zhou2017scene}) and classification (ImageNet-1K~\cite{ILSVRC15}). To support drop-in convolutions, we remove the [cls] token from inputs. For segmentation, this has no impact; for classification, we use the mean of tokens as decoder input, with negligible performance loss. Convolutional layers use a fixed kernel size $k=3$ for efficiency across all tasks.

\subsection{Convolution Attention Results}
\label{sec:convatt_results}
\subsubsection{Head-level Profiling.}
In \Cref{tab:head_benchmark}, we profile a single multi-head attention block, providing a straightforward setting to quantify performance differences. Both in the unensembled setup (lines 2-3) or in the ensembled one (lines 4-5), it is clear that the depthwise formulation is advantageous, speeding up execution by a factor $3\times$ with respect to the full convolution. The ensembled formulation is significantly faster, although, as discussed later in this chapter, it results in a more significant performance drop. Memory usage patterns are less intuitive; the separable formulation has a slightly higher memory requirement due to intermediate results and less effective buffer reuse, which is offset in the ensembled formulation.
\begin{table*}[ht!]

\centering
\caption{Comparison of different choices for the MhSA block. All results are measured for a single MhSA block with $n_h=16$ heads and an input size of $24\times24$ patches (Equivalent to an image resolution of $336\times\ 336$ for Dino-V2). Batch size is set as 1. }
    \begin{tabular}{lccccc} %
    \toprule[0.1em]
    \textbf{Attention} & \textbf{FLOPs (G)} & \textbf{Params (M)} & \textbf{Inference (ms)} & \textbf{Memory (MB)}\\
    \midrule
        MhSA (\cref{eq:mhsa})    & 6.19  & 4.2        & 3.2   & 47.2  \\
        \midrule
        Conv\textsubscript{\texttt{[all]}} (\cref{eq:attconv_full})    & 12.08 & 2.1        & 3.71  & 4.5   \\
        DW\textsubscript{\texttt{[all]}} (\cref{eq:attconv_dwpw})    & 2.43  & 2.11 & \textbf{1.26}  & 6.75  \\
        \midrule
        SPViT\textit{-style} (\cite{he2024pruning})      & 0.75  & 2.1  & 0.641 & 4.5   \\
        Ens+DW (\cref{eq:mhsa_dw_ensemble}) & 0.15 & 2.1          & \textbf{0.215} & 0.288 \\
    \bottomrule
    \end{tabular}
\label{tab:head_benchmark}
\end{table*}

\subsubsection{Full Model Performance.} 
\label{subsec:422}

When not otherwise specified, we first perform fine-tuning on the target task with regular MhSA heads, replace the selected heads with convolutional layers, and perform a new fine-tuning for half of the training epochs. The complete experimental setup is detailed in the Appendix. In \Cref{tab:bigtable} we compare the results obtained with different options for the MhSA block. For all experiments, we use the standard deviation criterion proposed in \Cref{subsec:stc_crit} with the blockwise selection. Despite not aiming for state-of-the-art performance, the strength of Dino-V2 features makes our baseline results highly competitive. Latency is reported only for the ViT backbone to avoid being affected by the design choices of the decoder.

\paragraph{Evaluation}
We first confirm that the depthwise formulation matches the performance of full convolution (\textit{CL2–CL3}), in line with our analytical derivation, while delivering a substantial speedup in inference. Without head ensembling, the full convolution baseline is 7.5\% slower than MhSA (\textit{CL1}), whereas the depthwise variant achieves a 17.2\% speedup. For a fair comparison with the ensembled setup, we match configurations of $|\mathcal{S}|$ with similar observed speedup $|\mathcal{S}|$: $12/24$ blocks for the unensembled case and $10/24$ for the ensembled case. In this setting, depthwise unensembled (\textit{CL3}) incurs a smaller accuracy drop (-0.08 vs. -0.98 mIoU) than the SPViT-style full-convolution ensemble, while still providing a modest speedup advantage. When combined with head ensembling (\textit{CL5}), the depthwise formulation achieves an 18.8\% speedup, but at the cost of a larger -1.6 mIoU drop. The same trade-off is observed when increasing $|\mathcal{S}|$ to 16/24 and 12/24 (\textit{CL6–CL7}). Comparable results are observed on the smaller ViT-B backbone (\textit{CB1–CB3}) and on the ImageNet classification task (\textit{IL1–IL5}), confirming the consistency of these observations across model scale and heterogeneous tasks. Additional results on ViT-B and on ADE20K dataset are reported in \Cref{tab:moreres} \\

\begin{table*}[h!]
\centering
\caption{Results on COCO and Imagenet with different formulations. Results are obtained finetuning Dino-V2, $336\times336$ input resolution. Inference performances are reported with batch-size=1.}
\begin{adjustbox}{width=1\textwidth}
    \setlength\tabcolsep{8pt} %
    \begin{tabular}{llllccccc} %

    \toprule[0.1em]
    \textbf{ID} & \textbf{Task} & \textbf{ViT} & \textbf{Attention} & $\bm{|\mathcal{S_{}}|}$ & \textbf{mIoU}  & \textbf{$\delta$-mIoU}  & \textbf{Infer (ms)} & \textbf{Speedup (\%)} \\
    \midrule
    
    \textit{CL1} & \multirow{11}{*}{\textbf{COCO}}   & \multirow{7}{*}{\textbf{Large}}   & \textbf{MhSA}    & -        & 66.03  & (baseline)   & 161.4    & (baseline)  \\
    \cmidrule{4-9}
    \textit{CL2} & & & Conv\sub{all}    & 12/24    & 65.84 & -0.19       & 173.5  & -7.49 \\
    \textit{CL3} & & & DW\sub{all}        & 12/24    & 65.95 & -0.08      & 133.6     & 17.21          \\
    \textit{CL4} & & & SPViT\textit{-style}   & 10/24    & 65.05 & -0.98   & 133.3     & 17.40          \\
    \textit{CL5} & & & Ens + DW & 10/24    & 64.81 & -1.61   & 131.1     & 18.79          \\
    \cmidrule{4-9}
    \textit{CL6} & & & DW\sub{all}       & 16/24    & 64.64 & -1.39   & 126.1     & 21.85          \\
    \textit{CL7} & & & Ens + DW& 12/24    & 63.92 & -2.11   & 124.5     & 22.87          \\

    \cmidrule[0.1em]{3-9}
    \textit{CB1} && \multirow{3}{*}{\textbf{Base}} & MhSA      & -    & 63.37 & (baseline)      & 49.9   & (baseline)  \\
    \textit{CB2} &&  & DW\sub{all}      & 6/12 & 62.22 & -1.16 & 40.8 & 18.10 \\
    \textit{CB3} &&  & Ens + DW & 6/12 & 60.55      &  -2.83      & 38.1 & 23.52 \\

    \toprule[0.1em]
    \textbf{ID} & \textbf{Task} &  \textbf{ViT} & \textbf{Attention} & $\bm{|\mathcal{S_{}}|}$ & \textbf{Top-1 Acc.}  &  \textbf{$\delta$-Acc} & \textbf{Infer (ms)} & \textbf{Speedup (\%)} \\
    \midrule
    
    \textit{IL1} & \multirow{6}{*}{\textbf{Imagenet}} & \multirow{6}{*}{\textbf{Large}} & \textbf{MhSA}     & -        & 86.22  & (baseline)    & 161.4     & (baseline)           \\
    \cmidrule{4-9}
    \textit{IL2} & &  & DW\sub{all}          & 12/24       & 85.45 & -0.77      & 133.6     & 17.21          \\
    \textit{IL3} & &  & Ens + DW& 10/24    & 84.96 & -1.26    & 131.1     & 18.79          \\
    \cmidrule{4-9}
    \textit{IL4} & &  & DW\sub{all}       & 16/24    & 84.88  & -1.34     & 126.1     & 21.85          \\
    \textit{IL5} & &  & Ens + DW& 12/24    & 84.65 &  -1.57  & 124.5     & 22.87         \\
    \bottomrule[0.1em]
    \end{tabular}
\end{adjustbox}
\label{tab:bigtable}
\end{table*}

\noindent \textbf{Generalization.}
We evaluate the generalization of our approach to backbone models beyond Dino-V2 by considering CLIP (\cite{radford2021learning}) and MAE (\cite{he2022masked}). Despite their different pretraining objectives, both models share the same underlying ViT architecture, allowing us to replicate our training setup without modification. For each model, we evaluate both the ViT-Base and ViT-Large variants, with patch sizes indicated by /16 and /14, respectively. Results are reported in \Cref{tab:dinoExtention}. We use the unensembled formulation with blockwise selection, applying the $\Sigma_b$ criterion to identify the block set. Observing the drop in mIoU relative to the same backbone without drop-in convolutions (i.e., $|\mathcal{S}| = -$), we observe results that mirror those obtained with Dino-V2, further validating the generality of our drop-in formulation across different vision foundation models.

\begin{table}[h!]
    \centering
    \begin{minipage}[b]{0.48\columnwidth}
        \centering
        
\caption{Results on different tasks using the Depthwise formulation.}
\adjustbox{max width=\linewidth}{
    \centering
    \begin{tabular}{lcccc} %
    \toprule[0.1em]
    \textbf{Task} & \textbf{ViT} & \textbf{Attention} & $\bm{|\mathcal{S_{}}|}$ & \textbf{Acc.}\\
    \midrule
        \multirow{2}{*}{\textbf{Imagenet}} & \multirow{2}{*}{\textbf{Base}}  & MhSA & -     & 83.76  \\
         &   & DW\sub{all} & 6/12  & 82.00  \\
        \midrule
        \textbf{Task} & \textbf{ViT} & \textbf{Attention} & $\bm{|\mathcal{S_{}}|}$ & \textbf{mIoU}\\
        \midrule
        \multirow{4}{*}{\textbf{ADE20K}}   & \multirow{2}{*}{\textbf{Base}}  & MhSA & -     & 53.83 \\
       &   & DW\sub{all} & 6/12  & 51.70 \\
       \cmidrule{2-5}
       & \multirow{2}{*}{\textbf{Large}} & MhSA & -     & 56.84 \\
       &  & DW\sub{all} & 12/24 & 56.05 \\
    \bottomrule
\end{tabular}
}
\label{tab:moreres}

    \end{minipage}\hfill
    \begin{minipage}[b]{0.48\columnwidth}
        \centering
        
\caption{Evaluation of Depthwise formulation applied to MAE and CLIP fine-tuned on COCO Semantic segmentation.}
\adjustbox{max width=\linewidth}{
\begin{tabular}{ccccc}
\toprule[0.1em]

\textbf{Model}                & \textbf{Backbone}             &  $\bm{|\mathcal{S_{}}|}$                & \textbf{mIoU} & \textbf{Infer (ms)}
\\
\midrule
MAE                  & ViT-B/16             & -                   & 58.84   & 42.95            \\
MAE                  & ViT-B/16             & 6/12               & 57.91   & 35.04          \\
MAE                  & ViT-L/16             & -                   & 60.22       & 122.47        \\
MAE                  & ViT-L/16             & 12/24                 & 59.81   & 102.23        \\
\midrule
CLIP                 & ViT-B/16             & -                   & 61.52    & 42.93            \\
CLIP                 & ViT-B/16             & 6/12               & 59.06   & 35.07        \\
CLIP                 & ViT-L/14             & -                   & 64.65   & 153.91            \\
CLIP                 & ViT-L/14             & 12/24                 & 62.17  & 128.38 \\      
\bottomrule[0.1em]
\end{tabular}
}
\label{tab:dinoExtention}

    \end{minipage}
\end{table}
\subsection{Selection Mechanism Results}

\label{subsec:results_selection}

\begin{table}[h!]
    \centering
        \caption{Comparison of selection mechanisms.}
    \begin{subtable}[t]{0.465\columnwidth}
        \centering
        \adjustbox{max width=\linewidth, max height=0.1\textheight}{
        \begin{tabular}{lcccc}
            \toprule[0.1em]
            \textbf{Crit.} & \textbf{Stages} & \textbf{BW} & $\bm{|\mathcal{S}|}$ & \textbf{mIoU}\\
            \midrule
            $\Sigma_{b}$\sub{LOWEST} & 2 & \multirow{3}{*}{Yes} & 12/24 & 65.95 \\
            DSP - e2e    & 2 & & 12/24 & 64.15 \\
            DSP - 2S     & \textbf{3} & & 12/24 & 64.00 \\
            \midrule
            $\Sigma_{b}$\sub{LOWEST} & 2 & \multirow{3}{*}{Yes} & 17/24 & 63.77 \\
            DSP - e2e    & 2 & & 17/24 & 58.25 \\
            DSP - 2S     & \textbf{3} & & 17/24 & 64.20 \\
            \bottomrule[0.1em]     
            \end{tabular}
            }
        \caption{Comparision between $\Sigma_b$ and DSP in Blockwise setting}
        \label{tab:select:blockwise}
    \end{subtable}\hfill
    \begin{subtable}[t]{0.5\columnwidth}
        \centering
        \adjustbox{max width=\linewidth}{
            \begin{tabular}{lcccc}
            \toprule[0.1em]
            \textbf{Crit.} & \textbf{Stages} & \textbf{BW} & $\bm{|\mathcal{S}|}$ & \textbf{mIoU}\\
            \midrule
            $\Sigma_{h}$\sub{LOWEST} & 2 & \multirow{3}{*}{No} & 192/384 & 65.52 \\
            DSP - e2e    & 2 & & 192/384 & 62.32 \\
            DSP - 2S     & \textbf{3} & & 192/384 & 65.79 \\
            \midrule
            $\Sigma_{b}$\sub{\textbf{HIGHEST}} & 2 & \multirow{3}{*}{Yes} & 12/24 & 61.46 \\
            $\Sigma_{b}$\sub{\textbf{HIGHEST}} & 2 & & 7/24 & 64.02 \\
            $\Sigma_{b}$\sub{LOWEST} & \textbf{1} & & 12/24 & 65.63 \\
            \bottomrule[0.1em]     
            \end{tabular}
        }
        \caption{(top) Comparision between $\Sigma_h$ and DSP in Scattered setting (bottom) Ablation of $\Sigma_b$: selection of the worst candidates ( $\Sigma_{b}$\sub{\textbf{HIGHEST}}) and 1-stage finetuning.}
        \label{tab:select:other}
    \end{subtable}
    \label{tab:select}
\end{table}

\begin{figure}[h!]
    \centering
    \begin{minipage}[t]{0.48\textwidth}
        \centering
        \includegraphics[width=\linewidth, trim={0 10 0 0},clip]{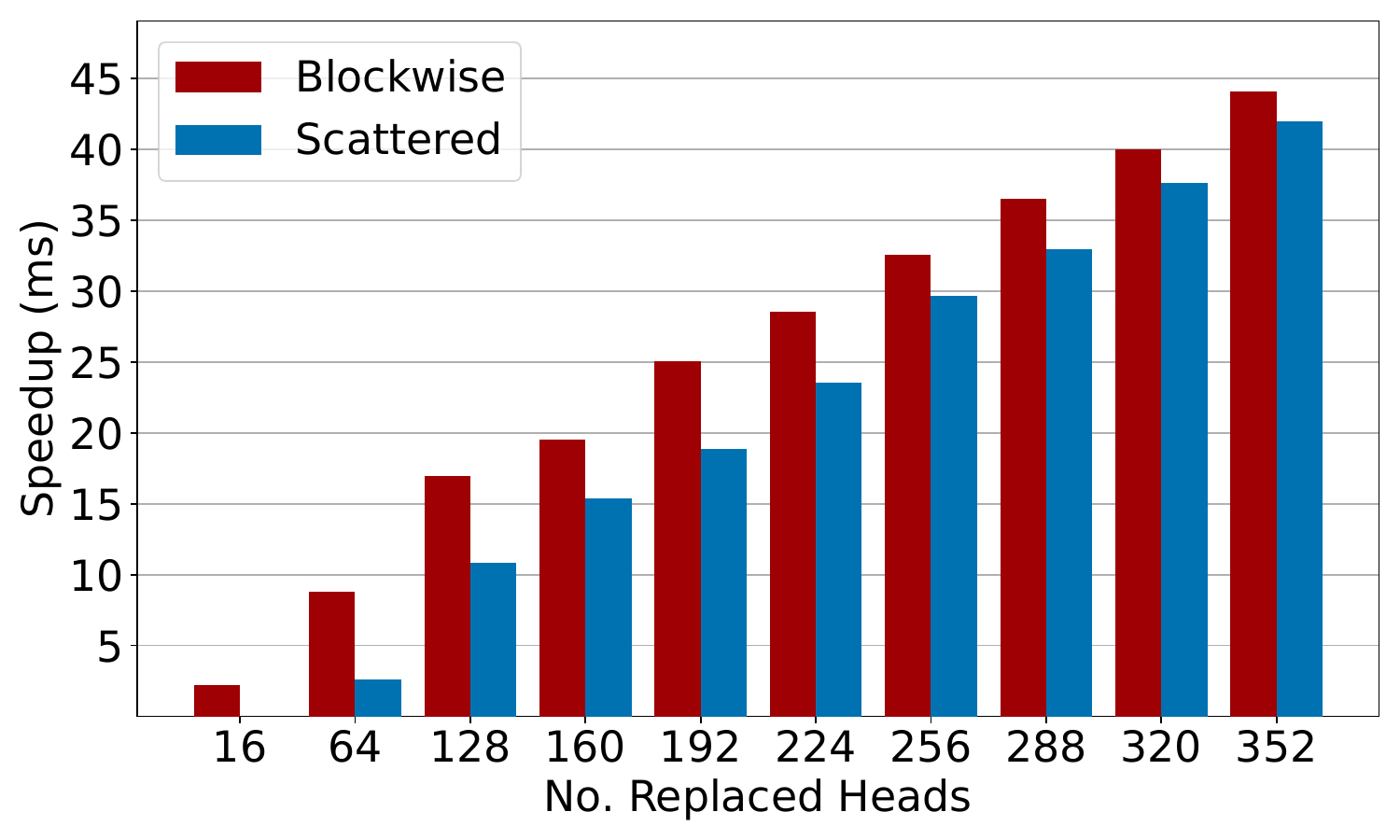}
        \caption{Speedup vs number of heads replaced in blockwise and scattered setups. Results on ViT-L (24 blocks, 16 heads per block $336\times 336$).}
        \label{fig:const_vs_unconst}
    \end{minipage}\hfill
    \begin{minipage}[t]{0.48\textwidth}
        \centering
        \includegraphics[width=\linewidth,clip]{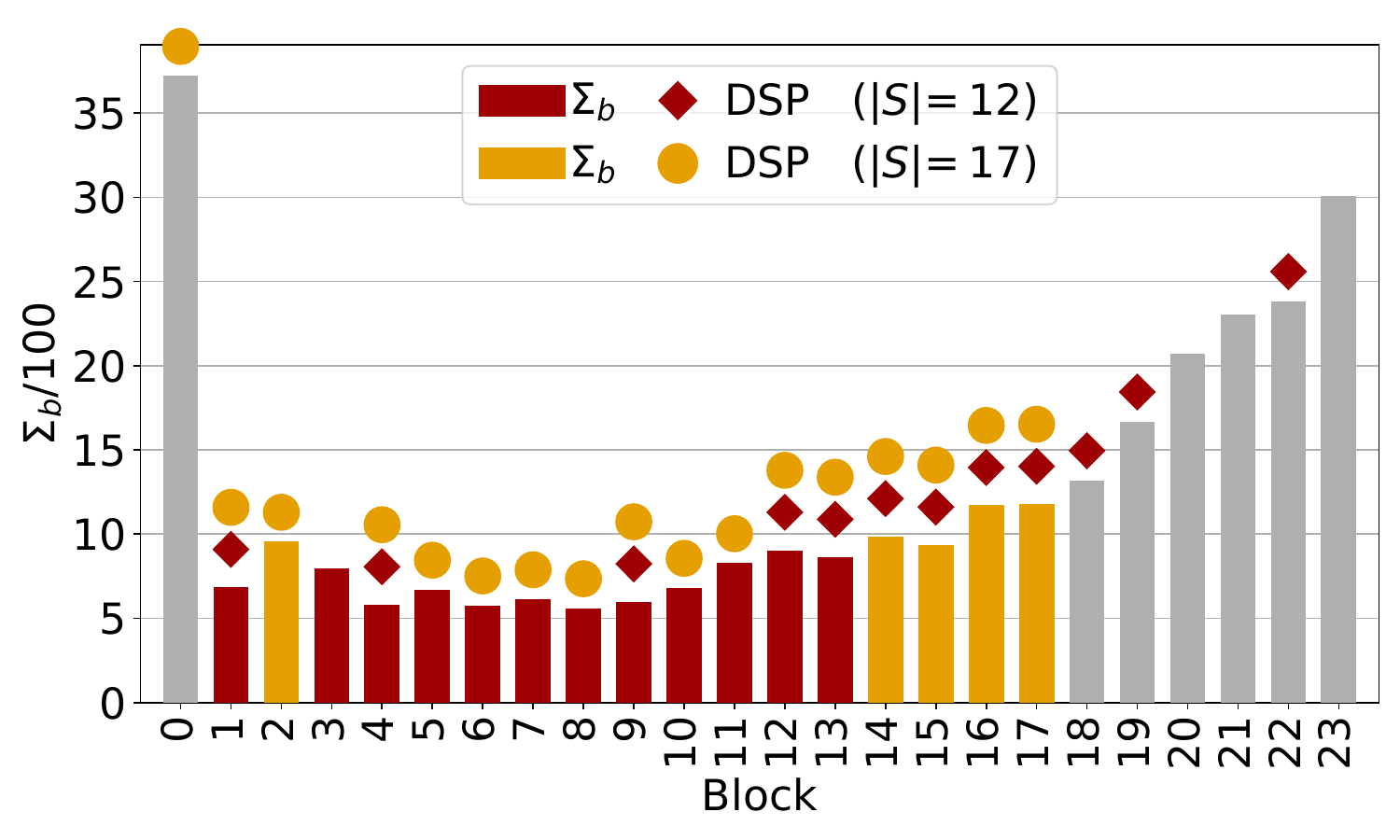}
        \caption{Blocks selected using DSP and $\Sigma_b$ criterion in blockwise setting. } %
        \label{fig:selblocks}
    \end{minipage}
\end{figure}

We first discuss the implications of blockwise and scattered selection. While for blockwise selection the exact subset $\mathcal{S}_b^{[p_b]}$ has no impact on the observed speedup, in the scattered setting the distribution of $\mathcal{S}_h^{[p_h]}$ is relevant. Scattered heads selection causes speedup to scale non-linearly with the number of selected heads due to overhead from memory and multiple kernels execution, which can offset gains. The impact on model performance is clearly observable in \Cref{fig:const_vs_unconst}, with the blockwise selection being consistently faster, while only implying a small performance drop (\Cref{tab:select}). 

\noindent \textbf{Comparison.} In \Cref{tab:select} we compare the proposed selection heuristic (\Cref{subsec:sigmacrit}), with differentiable subset pruning (DSP) \Cref{subsec:dsp}. For the latter, we evaluate both end-to-end training of selection gates (DSP-e2e, \Cref{eq:gumbel_mhsa}) and a two-stage variation (DSP-2S), where the learned set is reused in a new fine-tuning run with fixed selection, discarding the end-to-end weights. In the blockwise setup (\Cref{tab:select:blockwise}) with $|\mathcal{S}|=12/24$ the $\Sigma_b$ criterion outperforms the DSP, and DSP-2S reduces the gap only slightly, still trailing by nearly 2 mIoU points. A closer inspection (\Cref{fig:selblocks}), reveals that DSP often selects high-variance heads. Increasing to $|\mathcal{S}|=17/24$ DSP-2S slightly surpasses $\Sigma_b$, while DSP-e2e suffers from severe training instabilities and degraded performance.

\noindent\textbf{Ablation.} To further test our criterion, we also evaluate replacing the worst heads (highest $\Sigma_{E^{h}}$) in \Cref{tab:select:other}. With only 7 blocks replaced, the performance drop already exceeds that of the best 12 blocks, and replacing the 12 worst blocks leads to severe degradation. Finally, we ablate the fine-tuning procedure (last row): a single-stage fine-tuning, where convolutions are applied directly, achieves performance close to the two-stage setup.

\section{Conclusions and Future Work}
\label{sec:conc}
We assessed the effectiveness of a simple drop-in replacement for attention heads exhibiting convolution-like behavior in large-scale pretrained ViTs.
The proposed framework achieved 17\% speedup with minimal impact on downstream performance, highlighting that many pretrained heads can be approximated by efficient depthwise convolution without losing their functional role and thus largely preserving the power of pretrained weights. The investigated approach is not an alternative to existing pruning techniques, but rather as a component that can also be effective in combination with existing solutions. %

\subsubsection{Acknowledgements} This research was partially funded by the dAIedge project (HORIZON-CL4-2022-HUMAN-02-02, Grant Agreement Number: 101120726) and the Ministry of Education and Science of Bulgaria (support for INSAIT, part of the Bulgarian National Roadmap for Research Infrastructure).

\bibliographystyle{splncs04}
\bibliography{main}

\appendix
\renewcommand{\theequation}{A.\arabic{equation}}
\renewcommand{\thefigure}{A.\arabic{figure}}
\renewcommand{\thetable}{A.\arabic{table}}
\renewcommand{\thesection}{A.\arabic{section}}
\setcounter{equation}{0}
\setcounter{figure}{0}
\setcounter{table}{0}
\end{document}